\documentclass[lettersize,journal]{IEEEtran}
\usepackage{amsmath,amsfonts}
\usepackage{algorithmic}
\usepackage{algorithm}
\usepackage{array}
\usepackage[caption=false,font=normalsize,labelfont=sf,textfont=sf]{subfig}
\usepackage{textcomp}
\usepackage{stfloats}
\usepackage{url}
\usepackage{verbatim}
\usepackage{graphicx}
\usepackage{cite}
\usepackage[table]{xcolor}
\usepackage{multirow}
\usepackage{graphicx}
\usepackage{booktabs}
\usepackage{amsmath}
\usepackage{bm}
\usepackage{mathtools}
\usepackage{pifont}
\usepackage[export]{adjustbox}
\usepackage{hyperref}
\definecolor{gray}{gray}{0.5}
\definecolor{myLightBlue}{RGB}{220, 235, 250}

\newcommand{\eg}[1]{\textit{e.g.,}}
\newcommand{\ie}[1]{\textit{i.e.,}}
\newcommand{\pub}[1]{{\color{gray}{\tiny{[{#1}]}}}}
\newcommand{\xhdr}[1]{\vspace{1.7mm}\noindent{{\bf #1.}}}

\begin{document}

\title{Learning to Fuse and Reconstruct Multi-View Graphs for Diabetic Retinopathy Grading}

\author{Haoran Li, \IEEEmembership{Student Member, IEEE}, Yuxin Lin, Huan Wang, \IEEEmembership{Student Member, IEEE}, Xiaoling Luo, Qi Zhu, Jiahua Shi, Huaming Chen, \IEEEmembership{Member, IEEE}, Bo Du, Johan Barthelemy, Zongyan Xue, Jun Shen, \IEEEmembership{Senior Member, IEEE}, and Yong Xu, \IEEEmembership{Senior Member, IEEE}
\thanks{H. Li is with the Department of Data Science and Artificial Intelligence, Monash University, Melbourne, Australia. H. Li is also with the ARC Centre of Excellence for the Weather of the 21st Century}
\thanks{H. Wang and J. Shen are with the School of Computing and Information Technology, University of Wollongong, Wollongong, Australia.}
\thanks{Y. Lin and Y. Xu are with the Shenzhen Key Laboratory of Visual Object Detection and Recognition, Harbin Institute of Technology (Shenzhen), Shenzhen, China.}
\thanks{X. Luo is with the College of Computer Science and Software Engineering, Shenzhen University, Shenzhen, China}
\thanks{Q. Zhu is with the College of Artificial Intelligence, Nanjing University of Aeronautics and Astronautics, Nanjing, China}
\thanks{J. Shi is with the Centre for Nutrition and Food Sciences, The University of Queensland, Brisbane, Australia}
\thanks{H. Chen is with the School of Electrical and Computer Engineering, University of Sydney, Sydney, Australia}
\thanks{B. Du is with the Department of Management, Griffith University, Brisbane, Australia}
\thanks{J. Barthelemy is with the NVIDIA, Santa Clara, USA}
\thanks{Z. Xue is with The University of New South Wales, Sydney, Australia}
\thanks{Corresponding author: Jun Shen, email: jshen@uow.edu.au and Yong Xu, email: laterfall@hit.edu.cn}}

\markboth{arxiv}%
{Shell \MakeLowercase{\textit{et al.}}: A Sample Article Using IEEEtran.cls for IEEE Journals}


\maketitle

\begin{abstract}
Diabetic retinopathy (DR) is one of the leading causes of vision loss worldwide, making early and accurate DR grading critical for timely intervention. Recent clinical practices leverage multi-view fundus images for DR detection with a wide coverage of the field of view (FOV), motivating deep learning methods to explore the potential of multi-view learning for DR grading. However, existing methods often overlook the inter-view correlations when fusing multi-view fundus images, failing to fully exploit the inherent consistency across views originating from the same patient. In this work, we present MVGFDR, an end-to-end \textbf{M}ulti-\textbf{V}iew \textbf{G}raph \textbf{F}usion framework for \textbf{DR} grading. Different from existing methods that directly fuse visual features from multiple views, MVGFDR is equipped with a novel Multi-View Graph Fusion (MVGF) module to explicitly disentangle the shared and view-specific visual features. Specifically, MVGF comprises three key components: \ding{172} Multi-view Graph Initialization, which constructs visual graphs via residual-guided connections and employs Discrete Cosine Transform (DCT) coefficients as frequency-domain anchors; \ding{173} Multi-view Graph Fusion, which integrates selective nodes across multi-view graphs based on frequency-domain relevance to capture complementary view-specific information; and \ding{174} Masked Cross-view Reconstruction, which leverages masked reconstruction of shared information across views to facilitate view-invariant representation learning. Extensive experimental results on MFIDDR, by far the largest multi-view fundus image dataset, demonstrate the superiority of our proposed approach over existing state-of-the-art approaches in diabetic retinopathy grading.
\end{abstract}

\begin{IEEEkeywords}
Diabetic Retinopathy Detection, Visual Graph Learning, Feature Fusion, Medical Imaging, Image Classification.
\end{IEEEkeywords}

\section{Introduction}\label{sec:intro}
\IEEEPARstart{A}{s} one of the most prevalent complications of diabetes, diabetic retinopathy (DR) is recognized as a leading cause of blindness~\cite{oh2021early,hu2017inhibition}. DR is a microvascular complication that often causes damage to small blood vessels, including leakage, occlusion, and the formation of abnormal new vessels, ultimately leading to impaired vision. 
Following the international DR Grading standard~\cite{world2006prevention,cleland2023comparing}, the diagnosis of DR is based on a series of lesions and can be divided into five levels from light to serious (grade 0-4): normal, mild, moderate, severe, and Proliferative Diabetic Retinopathy (PDR). Therefore, DR grading and detection are crucial for the early diagnosis of diabetic retinopathy, aiming to prevent disease progression and subsequent vision loss.
\begin{figure*}[!t]
    \centering
    \includegraphics[width=1\linewidth]{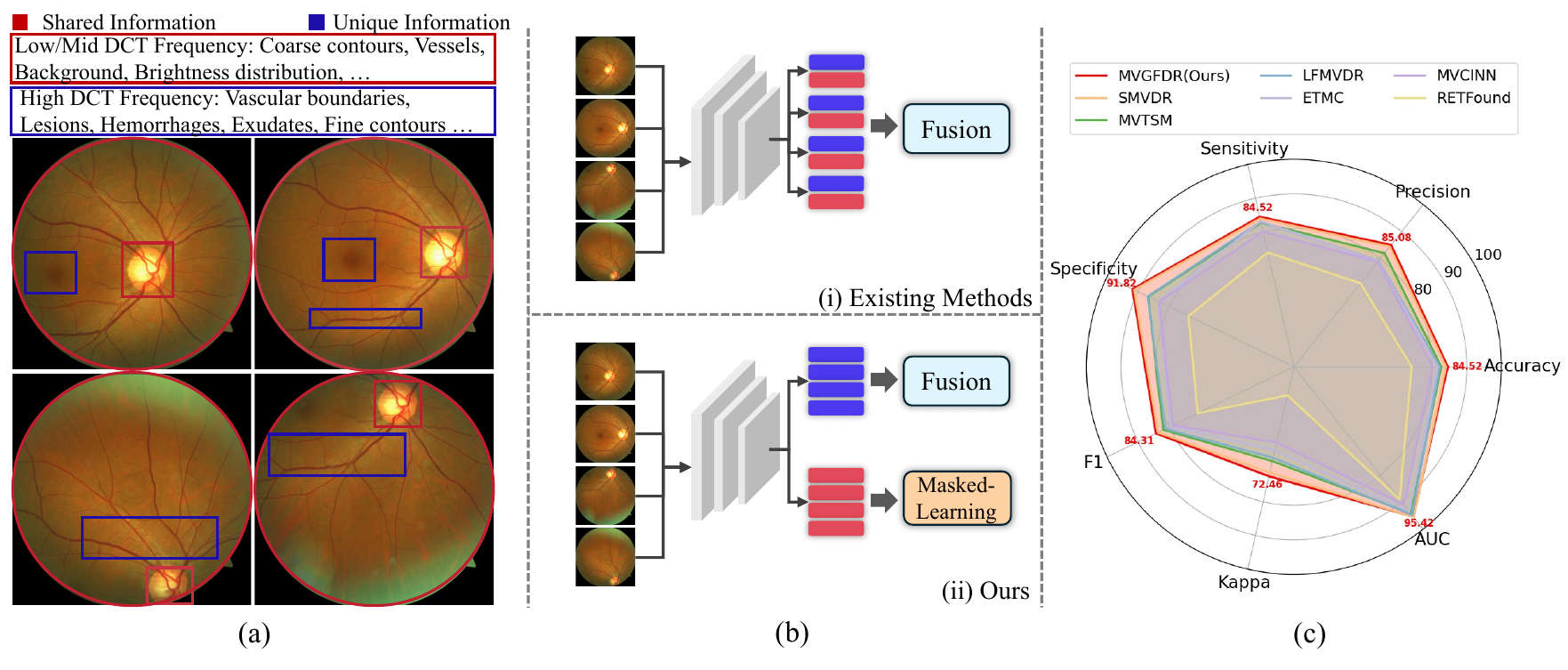}
    \caption{\textbf{Illustration of our motivation.} (a) Semantic information captured by different DCT frequency components in fundus images. (b) Comparison between (i) existing multi-view methods and (ii) our approach. Unlike prior methods, our approach disentangles the feature embeddings into shared and unique components, applying feature fusion and masked reconstruction learning separately to each. (c) Comparison of state-of-the-art multi-view DR methods and our proposed MVGFDR on seven evaluation metrics.}
    \label{fig:intro}
\end{figure*}

DR grading aims to assess the severity of diabetic retinopathy by identifying morphological lesions in fundus images. With the development of deep learning techniques, traditional expert-based visual examination is increasingly being supplanted by automated algorithms powered by computer vision~\cite{galappaththige2024generalizing,zhou2020benchmark}. Existing DR grading methods attempt to grade DR severity based on morphological analysis of fundus images~\cite{nguyen2020diabetic,nandy2021distributional}. Although they have made some progress, these methods still fall short of achieving high accuracy. The main reason is that most of the existing methods are still trained on single-view datasets with a field of view (FOV) of $45^{\circ}$ - $50^{\circ}$~\cite{porwal2020idrid,dai2024deep,pao2020detection}, unlike real clinical scenario, where physicians often examine the eye from a wider range of viewpoints, sometimes leveraging a three-dimensional perspective to assess the entire eyeball. A single-view-only FOV may lead to missing critical information, as certain lesions may only be visible from the alternative viewpoints. For example, as shown in Fig.~\ref{fig:intro}(a), a dark shadow is only visible from the top-left and top-right viewpoints.

To address this challenge, Luo et al.~\cite{luo2023mvcinn} proposed the first multi-view fundus image dataset for diabetic retinopathy detection. Building upon this, more recent studies have explored dual-stream frameworks that fuse information from different views to improve DR grading performance~\cite{luo2023mvcinn,lin2025multi,han2022trusted}. These approaches have demonstrated significant improvements over single-view methods by effectively incorporating complementary information. However, the aforementioned multi-view approaches often regard each input image as an isolated sample, neglecting their inherent correlations since all views originate from the same patient and are thus inherently correlated. As a result, they directly fuse visual features from multiple views without explicitly modeling inter-view relationships (as illustrated in Fig.~\ref{fig:intro}(b)(i)). These operations will lead to redundant information being processed, thereby degrading the overall performance of the model. As can be seen from Fig.~\ref{fig:intro}(a), fundus images captured from different viewpoints of the same patient typically exhibit shared morphologies, including coarse anatomical contours, global brightness patterns, primary vessel structures, etc. These common features may not benefit from further fusion. In contrast, the fusion should focus on view-specific cues, such as localized lesions and vascular boundaries, that offer complementary insights for accurate diagnosis.

From a medical imaging perspective, frequency-domain decomposition provides a principled way to separate shared anatomical structures from view-specific pathological details in multi-view fundus images, supported by extensive evidence from both ophthalmology and general medical imaging literature. Specifically, prior studies~\cite{g} have reported that pathological lesions in fundus images primarily manifest as fine-grained texture patterns, whereas background structures such as retinal vessels and the optic disc correspond to more stable anatomical configurations. In image signal analysis, high-frequency components are known to capture fine textures and local details, while low-frequency components mainly encode global background and structural information~\cite{h}. Moreover, in multi-view fundus images acquired from the same patient, clinically important anatomical indicators such as the cup-to-disc ratio~\cite{i} and vascular branching topology~\cite{j} remain largely invariant across viewpoints, suggesting that these shared anatomical characteristics are predominantly reflected in low-frequency representations. Beyond fundus imaging, similar observations have been reported in general medical imaging, where lesion-related signals tend to concentrate in high-frequency regions, while anatomical structures are associated with low-frequency components~\cite{k}. Collectively, these findings provide a principled medical and signal-processing rationale for leveraging frequency decomposition to distinguish shared anatomical information from view-specific pathological cues in multi-view fundus images.


Motivated by the above observations, we introduce a multi-view graph fusion (MVGF) strategy. As shown in Fig.~\ref{fig:intro}(b)(ii), our proposed MVGF aims to disentangle visual features into shared and view-specific components, and selectively fuse view-specific information, thereby reducing redundancy while enhancing discriminative representation. Moreover, to fully exploit the shared information, MVGF employs a masked graph learning strategy to reinforce robustness and cross-view consistency, further improving model generalization.

Specifically, based on this strategy, we propose a novel end-to-end \textbf{M}ulti-\textbf{V}iew \textbf{G}raph \textbf{F}usion framework for \textbf{DR} grading, termed MVGFDR. MVGFDR incorporates the above introduced MVGF module which comprises three key components: Multi-view Graph Initialization (MVGI), Multi-view Graph Fusion (MGF), and Masked Cross-View Reconstruction (MCVR). In MVGI, visual graphs are constructed using residual as guidance, with Discrete Cosine Transform (DCT) coefficients serving as cluster anchors. Feature disentanglement is achieved through node selection, leveraging the fact that different DCT anchors capture distinct frequency-domain information. In multi-view fundus images, shared and view-specific features often reside in different frequency bands (as illustrated in Fig.~\ref{fig:intro}(a)). MGF then takes the view-specific nodes from each visual graph and applies a Graph Convolutional Network (GCN) for feature fusion. To further improve robustness and cross-view understanding, MCVR takes shared nodes as input, masks a subset of nodes from one specific view, and reconstructs the missing information using the remaining three views.

In a nutshell, our contributions are in the follows:
\begin{itemize}
    \item[$\bullet$] We propose MVGFDR, a novel framework for multi-view DR grading. In contrast to prior works that utilize dual-stream structures to process multi-view information, our method employs a unified graph-based framework to jointly model inter-view relationships.
    \item[$\bullet$] MVGFDR integrates a multi-view graph initialization module and a graph fusion module to disentangle visual graphs from different views, enabling the selective fusion of unique, view-specific information to effectively reduce information redundancy.
    \item[$\bullet$] We further introduce a masked cross-view reconstruction module to enhance the model’s capability in multi-view understanding. By leveraging shared information across views, this module encourages the model to capture inter-view consistency and facilitates the learning of more robust and discriminative representations.
    \item[$\bullet$] Extensive experiments on MFIDDR, the largest publicly available multi-view fundus image dataset, together with two additional multi-view medical benchmarks and a newly constructed multi-view DR dataset, demonstrate that MVGFDR achieves state-of-the-art performance, consistently outperforming existing methods across multiple evaluation metrics.
\end{itemize}

The remainder of this paper is organized as follows. Section~\ref{sec:2} reviews related work. Section~\ref{sec:3} presents the proposed MVGFDR framework. Experimental results and analyses are reported in Section~\ref{sec:4}. Finally, Section~\ref{sec:5} concludes the paper.

\section{Related Work}\label{sec:2}
\subsection{Deep Learning in DR Grading}
Existing deep learning based DR Grading methods can be divided into two types: end-to-end DR Grading and expert-guided DR Grading.

\xhdr{End-to-end DR Grading} directly classify fundus images into severity levels. Gulshan et al.\cite{gulshan2016development} proposed a single-view CNN model to extract high-level semantics from local features. CABNet\cite{he2020cabnet} further integrates attention to highlight region-specific variations. Despite their success~\cite{yang2017lesion,araujo2020dr,li2022deep,ju2021synergic}, CNN-based models are limited in capturing long-range dependencies due to their constrained receptive fields. Hence, vision transformer-based methods~\cite{wu2021vision,zang2024cra,sadeghzadeh2023hybrid} are further explored to boost the performance of DR Grading. However, the inherent information bottleneck of single-view inputs makes it difficult for such methods to attain strong performance. CrossFiT~\cite{hou2022cross} proposes the first two-field DR Grading method, which fuses the fundus image information from two views through a cross-attention layer. MVCINN~\cite{luo2023mvcinn} introduces the first multi-view fundus dataset MFIDDR and proposes a hybrid framework that simultaneously aggregates both local and global information.

\xhdr{Expert-guided DR Grading} leverages additional information provided by experts as guidance to enhance the precision of DR classification through joint modeling with image data~\cite{sun2021lesion,huang2024ssit}. 
LFMVDR~\cite{luo2024lesion} enhances input images using lesion information, while SMVDR~\cite{Luo_Xu_Wu_Liu_Lai_Shen_2025} incorporates expert knowledge via visual prompts to guide attention. However, these methods treat each fundus view independently and fail to model inter-view correlations. In contrast, we propose an end-to-end DR grading framework based on graph fusion and reconstruction, which captures both shared and view-specific information. Notably, our MVGFDR outperforms existing methods even without relying on additional expert inputs.

\subsection{Visual Graph Learning}
Recent approaches on graph learning~\cite{yang2023simple,koh2024physicochemical} have proved their potential in mining complex relations among various scenarios. The application of graph-based techniques in vision models can be broadly categorized into two paradigms: pixel-as-graph~\cite{hu2020class,li2023zero,lin2021multilabel} and feature-as-graph representations~\cite{han2022vision,li2020spatial}. Owing to the flexibility of graph structures in representing visual embeddings, and the effectiveness of Graph Convolutional Networks (GCNs)\cite{kipf2017semisupervised} and Graph Attention Networks (GATs)\cite{veličković2018graph} in aggregating neighborhood information based on feature similarity or clustering patterns, graph-based fusion strategies have been widely adopted in traditional multi-view clustering tasks~\cite{wu2024ebmgc,ren2024dynamic}. Furthermore, several recent studies have attempted to apply graph fusion strategies to multi-modal data~\cite{wu2021scenegraphfusion,ding2025learnable}. Inspired by masked modeling techniques in vision and NLP, recent studies have further extended this paradigm to graph data~\cite{li2023s,hou2022graphmae}, showing that reconstructing masked nodes can significantly enhance graph-level representation learning. Motivated by these advances, we introduce a novel graph construction strategy based on DCT, which enables frequency-aware structural modeling across views. Specifically, we leverage DCT in the frequency domain to guide node selection across fundus image graphs. This enables the fusion of view-specific nodes that capture unique information while facilitating cross-view reconstruction among nodes that share similar frequency-aware semantics.

\section{Method}\label{sec:3}

\begin{figure*}[!t]
    \centering
    \includegraphics[width=1\linewidth]{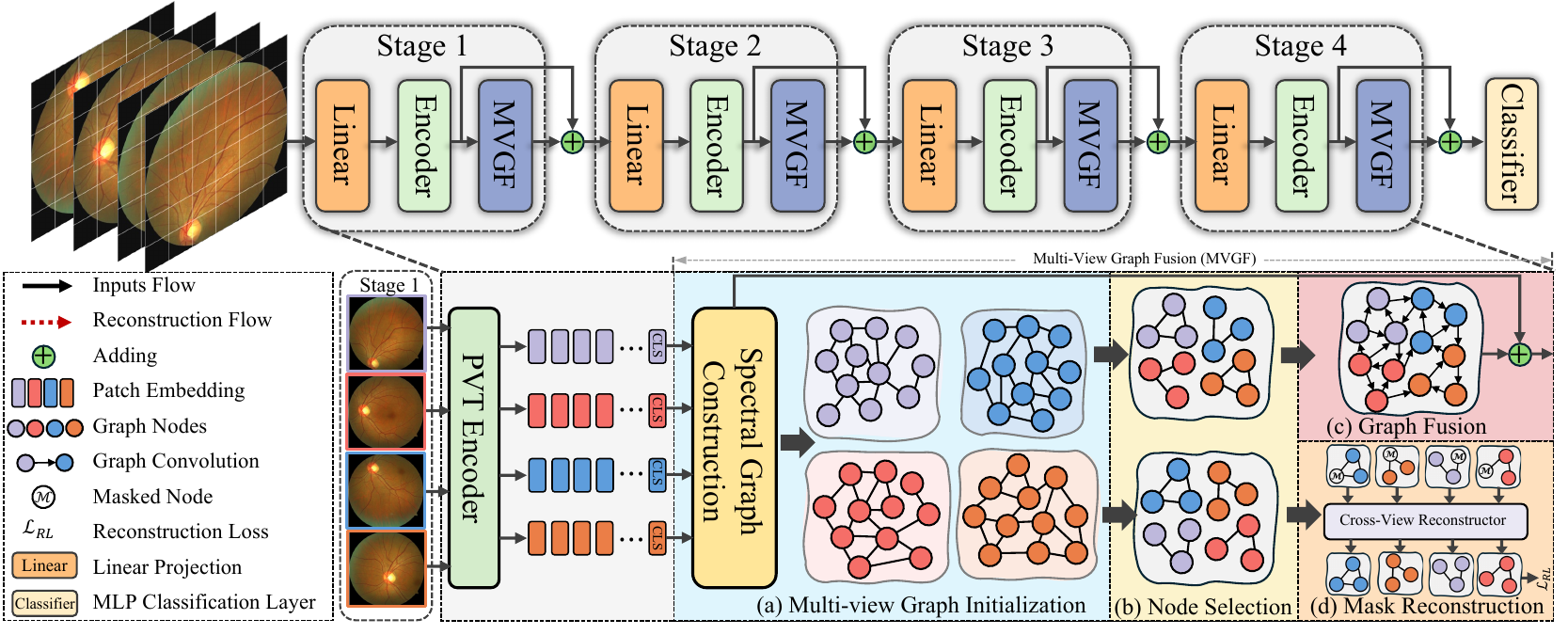}
    \caption{\textbf{Overall framework of MVGFDR.} (a) Multi-view Graph Initialization (MVGI) for each view based on DCT frequency. (b) Graph nodes selection from each view according to their corresponding high-, mid- and low-frequency DCT components. (c) Multi-view Graph Fusion (MGF) of high-frequency information using a GCN. (d) Masked Cross-View Reconstruction (MCVR) on low- and mid-frequency information with the random node masking rate $\eta$.}
    \label{fig:framework}
\end{figure*}

\subsection{Overview}
The overall architecture is shown in Fig.~\ref{fig:framework}. Given the input multi-view fundus images $X = [x_k]_{k=0}^{K}$, where $K$ denotes the number of views and $x_k \in \mathbb{R}^{H \times W \times 3}$, where $H$ and $W$ denote the height and width of the input fundus image.
Unlike existing multi-view DR grading methods~\cite{luo2023mvcinn,lin2025multi,Luo_Xu_Wu_Liu_Lai_Shen_2025} that utilize dual-stream architectures with parallel convolutional and transformer branches to capture both local and global receptive fields, our method adopts the Pyramid Vision Transformer (PVT)~\cite{wang2021pyramid} as a unified backbone that inherently integrates the strengths of CNNs and Transformers.
PVT adopts a four-stage architecture. At each stage, the input image is first divided into $N = \frac{HW}{hw}$ patches, with each patch having a size of $h \times w \times 3$. All the patches are further fed into the linear projection layer and transformer encoder to obtain the output features $F^{i}_k \in \mathbb{R}^{N_i \times C_i}$ at the $i$-th stage, where $C$ is the channel dimension. $F^{i}_k$ is then sent to the multi-view graph fusion (MVGF) module to obtain the fused multi-view features $\tilde{F}^{i}$ through the multi-view graph initialization (MVGI), node selection, and graph fusion. The non-selected graph nodes are further sent to the masked cross-view reconstruction (MCVR) module for masked cross-view learning. The features of each stage are updated through $F^{i}_k + \tilde{F}^{i}$, and resized to $\frac{H}{2^{i+1}} \times \frac{W}{2^{i+1}} \times C_i$ for the next stage. At the final stage, the [CLS] tokens from all views are concatenated and fed into a two-layer MLP classifier to produce the final prediction.

\subsection{Multi-View Graph Fusion}\label{sec:mvgf}
\xhdr{MVGI} Following~\cite{zhang2017deep,li2018beyond,zhai2021mutual}, the graph initialization is guided by residual information. Specifically, we apply a linear projection to transform the input feature $F_k^i$ into a graph node representation of size $C_i \times M$. Different from previous methods that randomly initialize all learnable anchors, we utilize DCT coefficients to initialize the anchor matrix $W \in \mathbb{R}^{M \times C_i}$. Hence, the clustering center for the $n$-th node can be formulated as 
\begin{equation}\label{eq:DCT}
    w_n = \sum_{c=0}^{C_i - 1}a_{c} \cdot cos[\frac{\pi}{C_i}(c + \frac{1}{2})n], \text{ } n\in [0, 1, ..., M],
\end{equation}
where $M$ is a hyper-parameter denoting the number of clustering centers, and $a_c$ denotes a learnable vector. Notably, although the DCT anchors are learnable, their frequency-domain structure remains unchanged during training (see appendix~\ref{app:dct_proof} for theoretical justification). The representation of each node is computed as
\begin{equation}\label{eq:GI}
    \hat{v}^{i}_{k,n} = \frac{\sum^{N_i}_{j=0}q_{k}^{j}(f^i_{k,j} - w_n)/\sigma_{n}}{\sum^{N_i}_{j=0}q_{k}^{j}}, \text{ } v^{i}_{k,n} = \frac{\hat{v}^{i}_{k,n}}{\|\hat{v}^{i}_{k,n}\|_2},
\end{equation}
where $\sigma_n$ represents the dimension-wise standard deviation estimated from the dataset, $f^i_{k,j}$ denotes the $j$-th feature vector of the $k$-th view visual feature $F^{i}_{k}$ from stage $i$, and $q^j_k$ denotes the soft-assignment 
\begin{equation}
    q_{k}^{j} = \frac{\text{exp}(-||(f^i_{k,j} - w_n)||^2/2\sigma_n^2)}{\sum_{n}\text{exp}(-||(f^i_{k,j} - w_n)||^2/2\sigma_n^2)}.
\end{equation}
Afterwards, we employ node selection on the multi-view graph nodes $\mathcal{V}^i_k = \{v^{i}_{k,n}\}^M_{n=0}$.

\begin{figure}[!t]
    \centering
    \includegraphics[width=1.0\linewidth]{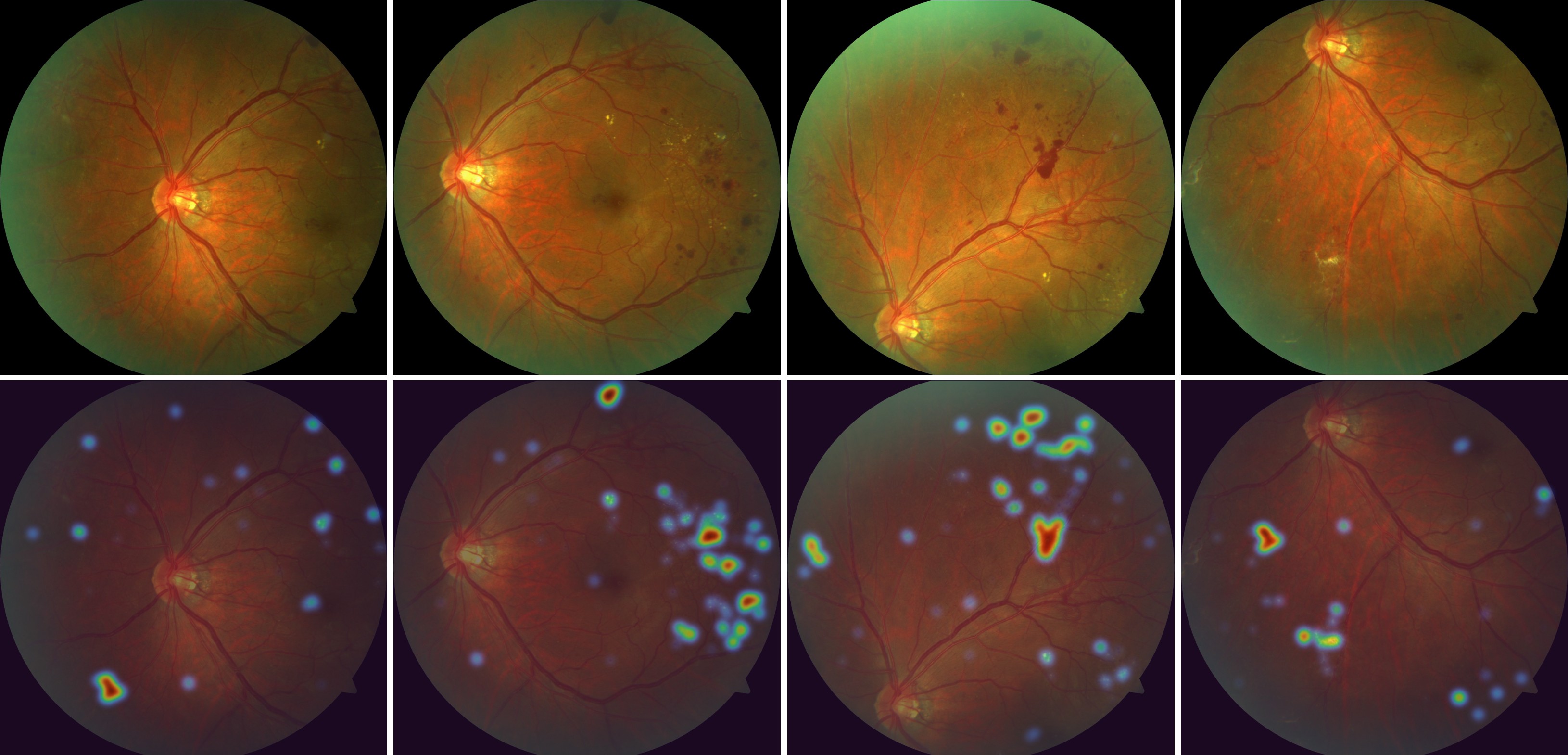}
    \caption{Visualization of view-specific representations extracted by MVGI from different views. A representative Grade-4 sample is shown for illustration. The visualization focuses on the high-frequency (DCT) representations, which are associated with lesion-related patterns.}
    \label{fig:feature_vis}
\end{figure}

\xhdr{Node Selection} To better fuse information from multiple views, different from existing multi-view methods that directly aggregate the features from different views, we decouple the visual graphs based on frequency-domain differences derived from DCT. Specifically, we separate each view's graph into shared-information nodes and view-specific nodes to facilitate selective fusion. Given the DCT clustering center from Eq.~\ref{eq:DCT}, we extract low-, mid-, and high-frequency nodes separately based on different index ranges
\begin{align}
&\mathcal{V}^{i}_{k_{{\text{(low+mid)}}}} = \{ v^{i}_{k,n} \mid 0 \leq n < k_{\theta}*M \}, \\
&\mathcal{V}^{i}_{k_{{\text{high}}}} = \{ v^{i}_{k,n} \mid k_{\theta}*M \leq n < M \},\\
&\mathcal{Q}^{i}_{k_{{\text{(low+mid)}}}} = \{ q^{j}_{k} \mid 0 \leq n < k_{\theta}*M \}, \\
&\mathcal{Q}^{i}_{k_{{\text{high}}}} = \{ q^{j}_{k} \mid k_{\theta}*M \leq n < M \},
\end{align}
where $k_{\theta}$ is a hyper-parameter defining frequency ranges for low-, mid-, and high-frequency components. As mentioned in Sec.~\ref{sec:intro}, because the multi-view fundus images are captured from the same patient, the low- and mid-frequency components (\ie, coarse contours, background, brightness distribution, and main vessels) across different views are expected to be consistent across views, while the high-frequency components (\ie, vascular boundaries, lesions, and exudates) typically capture view-specific information. To mitigate redundancy caused by overlapping information, we select high-frequency graph nodes from each view and fuse them via a graph fusion module to retain complementary details. In contrast, the low- and mid-frequency nodes are passed to the MCVR module for masked graph learning, enabling the model to better capture inter-view correlations and enhance its multi-view reasoning capability. An example of visualization of high-frequency features is provided in Fig.~\ref{fig:feature_vis}.

\xhdr{Graph Fusion} After obtaining the high-frequency nodes $\mathcal{V}_{k, \text{high}}^i$ and the soft-assignments $\mathcal{Q}_{k, \text{high}}^i$ from different views, we first combine all the nodes and their relevant soft assignments together
\begin{equation}\label{eq:combine}
    \mathcal{V}^i_{\text{high}} = \texttt{Concate}[\mathcal{V}_{k, \text{high}}^i]^K_{k=0}, \text{ } \mathcal{Q}^i_{\text{high}} = \texttt{Concate}[\mathcal{Q}_{k, \text{high}}^i]^K_{k=0},
\end{equation}
where $\texttt{Concate}[\cdot]$ denotes concatenation and $Q_{k, \text{high}}^i$ denotes the set of soft assignments corresponding to the high-frequency nodes. We further implement a graph convolution network (GCN) to fuse the unique information obtained from all views
\begin{equation}
    \mathcal{V}^{i}_{\text{fuse}} = g(\mathcal{W}_{f}^{(i)};\mathbf{A}\mathcal{V}^i_{\text{high}}), \text{ where } \mathbf{A} = \mathcal{N}(\mathcal{V}^{i\top}_{\text{high}} \cdot \mathcal{V}^i_{\text{high}}),
\end{equation}
where $g(\cdot)$ denotes a non-linear activation function, $\mathcal{W}_{f}^{(i)}$ denotes the learnable parameters of the GCN at the $i$-th stage, and $\mathcal{N}(\cdot)$ represents a normalization function. Finally, we utilize the merged soft assignments to reproject the fused graph into the feature space and update the original input features via a residual connection, yielding the input feature $F_{k,\text{input}}^{i+1}$ for the next stage
\begin{equation}
    F_{k,\text{input}}^{i+1} = (\mathcal{Q}^i_{\text{high}}\cdot\mathcal{V}^{i\top}_{\text{fuse}} + F_k^i) \in \mathbb{R}^{N_i\times C_i} \xrightarrow{{\text{reshape}}} \mathbb{R}^{\frac{H}{2^{i+1}} \times \frac{W}{2^{i+1}} \times C_i},
\end{equation}
where $F_k^i$ denotes the input feature of current stage.

\subsection{Masked Cross-View Reconstruction}
To enhance the model's ability to learn inter-view relationships and capture cross-view semantics, we introduce a Masked Cross-View Reconstruction (MCVR) module, inspired by the principles of masked image modeling (MIM)~\cite{he2022masked,xie2022simmim}. Unlike existing MIM methods, which focus on enhancing intra-image contextual understanding through self-supervised learning, MCVR is designed to leverage the knowledge learned from multiple views to reconstruct the information of the masked target view, enabling the model to reason over cross-view correlations more effectively. Hence, the MCVR is only applied to those graph nodes which share similar information from different views. In detail, given the low- and mid-frequency graph $\mathcal{V}^{i}_{k_{{\text{(low+mid)}}}}$, which share similar information across views, we first set one graph $\mathcal{V}_{K, \text{(low+mid)}}^i$ for masking and combine the remaining graphs following the same operation as Eq.~\ref{eq:combine}:
\begin{equation}
    \mathcal{V}_{\text{re}} = \texttt{Concate}[\mathcal{V}_{k, \text{(low+mid)}}^i]^{K-1}_{k=0}.
\end{equation}
Subsequently, we randomly perform masking over the nodes in $\mathcal{V}_{K, \text{(low+mid)}}^i$ 
\begin{equation}
    \mathcal{V}_{\text{masked}}^i \subset \mathcal{V}_{K, \text{(low+mid)}}^i, \text{ with } |\mathcal{V}_{\text{masked}}^i| = \eta \cdot |\mathcal{V}_{K, \text{(low+mid)}}^i|,
\end{equation}
where $\eta$ is a hyper-parameter that denotes the masking rate. Next, we perform masked node prediction by the proposed Cross-View Reconstructor. To better evaluate the effectiveness of our design, we additionally set a simple GCN-based reconstructor as a comparison baseline, 

\subsubsection{GCN-based Reconstructor (GCR)}\label{sec:gcr} 
\begin{figure}[!t]
    \centering
    \includegraphics[width=1\linewidth]{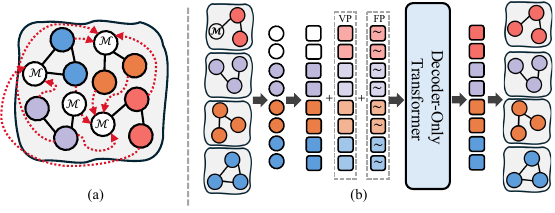}
    \caption{Proposed Masked Cross-View Reconstructor (MCVR) (a) We set a simple cross-graph reconstructor as the comparison baseline. (b) The proposed MCVR adopts a decoder-only transformer to perform masked view reconstruction guided by view-positional embeddings (VP, \protect\includegraphics[scale=1.0,valign=c]{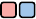}) and frequency-positional embeddings (FP, \protect\includegraphics[scale=1.0,valign=c]{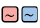}).}
    \label{fig:reconstructor}
\end{figure}

As illustrated in Fig.~\ref{fig:reconstructor}(a), we apply a GCN over the aggregated graph $\mathcal{V}_{\text{re}}$, followed by a two-layer Multi-Layer Perceptron (MLP) to produce the final predicted node features 
\begin{equation}
    \mathcal{V}_{\text{pred}} = \text{MLP}(g(\mathcal{W}_r^{(i)};\mathbf{A}\mathcal{V}_{\text{re}})), \text{ where } \mathbf{A} = \mathcal{N}(\mathcal{V}_{\text{re}}^\top \cdot \mathcal{V}_{\text{re}}),
\end{equation}
where $\mathcal{W}_{r}^{(i)}$ denotes the learnable parameters of the reconstruction GCN at the $i$-th stage. 

\subsubsection{Cross-View Reconstructor (CVR)}
Inspired by recent advances in generative large language models~\cite{radford2019language,dubey2024llama}, we adopt a decoder-only transformer as the base model to design a cross-view reconstructor. As shown in Fig.~\ref{fig:reconstructor}(b), CVR takes both aggregated graph $\mathcal{V}_{\text{re}}$ and the masked graph $\mathcal{V}_{\text{masked}}^i$ as the input. Since the graph nodes are already represented as feature embeddings, we directly flatten them into input tokens. To ensure that the reconstructed graph nodes preserve intra-view continuity while maintaining inter-view diversity, we design two different types of positional embeddings, view-positional embedding (VP) and frequency-positional embeddings (FP), to guide the reconstruction process.

\xhdr{VP and FP} The VP encodes the global view identity, ensuring that tokens from the same view share a unified positional prior
\begin{equation}
    P_{V} = [p_{k}]^{3}_{k=0},\quad p_k \in \mathbb{R}^{\text{dim}(v^i_{k,n})},
\end{equation}
where $p_{k}$ denotes a learnable positional embedding. We use the same VP within each view to ensure that the reconstruction process maintains view consistency and does not introduce ambiguity across views. However, within each view, the reconstruction process requires additional guidance to direct the generation of the CVR. To this end, we leverage the fact that the input nodes are distinguished according to their frequency-domain differences derived from the DCT, and introduce FP as an intra-view prior to ensure that the reconstruction is guided in the correct direction
\begin{equation}
    P_{F} = [e_{k,n}]^{3}_{k=0}, e_{k,n} = \omega \text{cos}(\frac{\pi n}{M}) + \rho \text{sin}(\frac{\pi n}{M}),
\end{equation}
where $\omega, \rho \in \mathbb{R}^{\text{dim}(v^i_{k,n})}$ denotes learnable parameters. Therefore, the input of the CVR could be represented as
\begin{equation}
\begin{split}
    \{v^i_{1,1}+p_{1}+e_{1,1},\dots,[\text{MSK}]^i_{k',n'}+&p_{k'}+e_{k',n'}, \dots, \\
    &v^i_{k,n}+p_{k}+e_{k,n}\},
\end{split}
\end{equation}
where $[\text{MSK}]^i_{k',n'}$ denotes the masked graph node. We further employ the commonly used decoder-only transformer~\cite{radford2019language,pang2025randar} as the reconstructor and feed the above tokens into it to obtain the final predicted node features $\mathcal{V}_{\text{pred}}$.

\xhdr{Target Function} Since the shared information across different views is similar but not strictly identical, we adopt a hybrid reconstruction loss to balance consistency and flexibility during training. Specifically, the reconstruction loss for view $k$ at stage $i$ is defined as
\begin{equation}
    \mathcal{L}^i_k = \mathcal{L}_{\text{cos}}(\mathcal{V}_{\text{pred}}^i,\mathcal{V}_{\text{masked}}^i) + \alpha \mathcal{L}_{\text{MSE}}(\mathcal{V}_{\text{pred}}^i,\mathcal{V}_{\text{masked}}^i),
\end{equation}
where $\alpha$ is a hyperparameter controlling the trade-off between cosine similarity loss $\mathcal{L}_{\text{cos}}$ and mean squared error (MSE) loss $\mathcal{L}_{\text{MSE}}$. And the reconstruction loss at each stage is computed by summing over all views $\mathcal{L}^i_{\text{recon}} = \sum^K_{k=0}\mathcal{L}^i_k$. Finally, the overall reconstruction loss is obtained by averaging over all stages.
Together with the classification loss, the overall objective function of the proposed MVGFDR framework is formulated as follows:
\begin{equation}
    \mathcal{L}_{\text{total}} = \mathcal{L}_{\text{cls}} + \bar{\mathcal{L}}_{\text{recon}},
\end{equation}
where $\mathcal{L}_{\text{cls}}$ denotes the focal loss~\cite{lin2017focal} used for DR grading and $\bar{\mathcal{L}}_{\text{recon}}$ denotes the average reconstruction loss for all the stages.

\begin{table*}[!t]
    \centering
    \caption{Comparison of precision (Prec.), sensitivity (Sens.), and F$_1$ scores between our method and single-view approaches in different DR grades. Best results are highlighted in bold.}
    \resizebox{\linewidth}{!}{
    \begin{tabular}{l |c c c| c c c|c c c|c c c|c c c}
        \hline
        & \multicolumn{3}{c}{Grade 0} & \multicolumn{3}{c}{Grade 1} & \multicolumn{3}{c}{Grade 2} & \multicolumn{3}{c}{Grade 3} & \multicolumn{3}{c}{Grade 4}\\
        \hline
         Method & Pre. & Sens. & F$_1$ & Pre. & Sens. & F$_1$ & Pre. & Sens. & F$_1$ & Pre. & Sens. & F$_1$ & Pre. & Sens. & F$_1$\\
         \hline
         Inception\_ResNet\_k\_2\_BSV & 74.83 & 97.15 & 84.54 & 46.28 & 12.53 & 19.72 & 48.98 & 39.34 & 43.64 & 63.38 & 60.81 & 62.07 & 66.67 & 15.38 & 25.00\\
         Mobile\_Net\_k2\_BSV & 76.60 & 96.70 & 85.49 & 48.55 & 14.99 & 22.91 & 56.85 & 45.36 & 50.46 & 64.16 & 75.00 & 69.16 & 60.00 & 15.38 & 24.49\\
         ResNet50\_BSV & 79.38 & 94.68 & 86.36 & 48.34 & 22.82 & 31.00 & 51.48 & 47.54 & 49.43 & 62.94 & 72.30 & 67.30 & 70.00 & 17.95 & 28.57\\
         ResNet50\_32x4d\_BSV & 79.51 & 95.13 & 86.62 & 46.64 & 27.96 & 34.97 & 55.04 & 38.80 & 45.51 & 71.23 & 70.27 & 70.75 & 75.00 & 23.08 & 35.29\\
         ConvNeXt-B\_BSV & 82.62 & 95.13 & 88.43 & 57.14 & 34.00 & 42.64 & 55.83 & 49.73 & 52.60 & 64.57 & \textbf{76.35} & 69.97 & 81.82 & 23.08 & 36.00\\
         Swin-B\_BSV & 81.03 & 95.43 & 87.64 & 54.36 & 29.31 & 38.08 & 54.02 & 51.37 & 52.66 & 68.87 & 70.27 & 69.57 & 92.86 & 33.33 & 49.06\\
         Vim\_BSV & 85.34 & 95.13 & 89.97 & 60.89 & 48.77 & 54.16 & 45.68 & 40.44 & 42.90 & 66.20 & 63.51 & 64.83 & \textbf{98.90} & 5.13 & 9.76\\
         PVT-M\_BSV & 80.72 & 97.90 & 88.48 & 54.24 & 28.64 & 37.48 & 50.54 & 51.37 & 50.95 & 72.28 & 49.32 & 58.63 & 70.00 & 17.95 & 28.57\\
         PVT-L\_BSV & 81.90 & \textbf{98.73} & 89.53 & 64.76 & 32.89 & 43.62 & 66.41 & 46.45 & 54.66 & 69.70 & 62.16 & 65.71 & 50.00 & \textbf{71.79} & 58.95\\
         RETFound\_BSV & 77.90 & 95.65 & 85.87 & 49.65 & 15.66 & 23.81 & 44.44 & 52.46 & 48.12 & 64.54 & 61.49 & 62.98 & 73.33 & 28.21 & 40.74\\
         \textbf{MVGFDR (Ours)} & \textbf{94.13} & 93.52 & \textbf{93.81} & \textbf{73.70} & \textbf{71.17} & \textbf{72.41} & \textbf{67.94} & \textbf{72.03} & \textbf{69.91} & \textbf{74.13} & 75.70 & \textbf{74.92} & 88.89 & 55.12 & \textbf{68.07}\\
         \hline
    \end{tabular}}
    \label{tab:per_result1}
\end{table*}

\begin{table}[!t]
    \centering
    \caption{Comparison results against single-view approaches with our proposed multi-view approach. Best results are in bold font.}
    \resizebox{\linewidth}{!}{
    \begin{tabular}{l c c c c c}
        \hline
         Method & Acc. & Spe. & Kappa & F$_1$ & AUC\\
         \hline
         Inception\_ResNet\_k\_2\_BSV~\pub{AAAI17}~\cite{szegedy2017inception} & 70.66 & 67.13 & 38.69 & 65.43 & 85.55\\
         Mobile\_Net\_k2\_BSV~\pub{CVPR18}~\cite{sandler2018mobilenetv2} & 72.38 & 68.73 & 43.61 & 67.28 & 86.88\\
         ResNet50\_BSV~\pub{CVPR16}~\cite{he2016deep} & 73.22 & 73.20 & 45.29 & 69.35 & 86.76\\
         ResNext50\_32x4d\_BSV~\pub{CVPR17}~\cite{xie2017aggregated} & 73.36 & 73.04 & 47.15 & 70.37 & 88.22\\
         ConvNeXt-B\_BSV~\pub{CVPR22}~\cite{liu2022convnet} & 75.96 & 77.81 & 53.72 & 73.65 & 89.77\\
         Swin-B\_BSV~\pub{CVPR21}~\cite{liu2021swin} & 75.08 & 75.53 & 51.32 & 72.42 & 88.83\\
         Vim\_BSV~\pub{ICML24}~\cite{vim} & 77.03 & 81.20 & 56.31 & 75.34 & 90.67\\
         PVT-M\_BSV~\pub{ICCV21}~\cite{wang2021pyramid} & 74.11 & 78.56 & 50.49 & 71.45 & 89.31\\
         PVT-L\_BSV~\pub{ICCV21}~\cite{wang2021pyramid} & 75.31 & 80.39 & 57.29 & 73.89 & 90.38\\
         RETFound\_BSV~\pub{Nature23}~\cite{zhou2023foundation} & 71.78 & 70.96 & 43.67 & 67.37 & 86.10\\
         \hline
         MVGFDR (CGR) & 84.52 & 91.82 & 72.46 & 84.31 & 95.42\\
         \rowcolor{myLightBlue}
         \textbf{MVGFDR (Ours)} & \textbf{85.87} & \textbf{93.00} & \textbf{73.14} & \textbf{84.49} & \textbf{96.17}\\
         \hline
    \end{tabular}}
    \label{tab:result1}
\end{table}

\section{Experiments}\label{sec:4}
\subsection{Experimental Setup}\label{sec:exp_setting}
\xhdr{Dataset} We conduct experiments on the largest publicly available multi-view diabetic retinopathy (DR) dataset, MFIDDR~\cite{luo2023mvcinn}. MFIDDR comprises $8,613$ eye samples, each associated with four fundus images (\ie, V1–V4) captured using a Zeiss Visucam NM/FA camera. Specifically, V1 focuses on the macula, V2 centers on the optic disc, while V3 and V4 are tangent to the upper and lower horizontal lines of the optic disc, respectively. This results in a total of $34,452$ high-resolution color fundus images with a size of $2124 \times 2556$ in total. All samples are annotated with DR grades following international standards. Following the experimental protocol of previous works~\cite{luo2023mvcinn,lin2025multi,luo2024lesion,Luo_Xu_Wu_Liu_Lai_Shen_2025}, the dataset is split into a training set and a testing set, containing $25,848$ and $8,604$ images, respectively. All the images are resized to $224 \times 224$ before training and testing.

\xhdr{Implementation details} We utilize the PVT~\cite{wang2021pyramid} pretrained on ImageNet as the backbone for our proposed MVGFDR. All experiments are conducted on a single NVIDIA A100 GPU (80GB) with a batch size of $40$. The model is trained for 100 epochs using Adam~\cite{kingma2014adam} with an initial learning rate of 1e-5. The number of clustering centers $M$ and frequency range $k_{\theta}$ are set to $32$ and $0.75$. And loss control rate $\alpha$ and masking rate $\eta$ for MCVR are set to $0.3$ and $0.5$, respectively. Please refer to Sec.~\ref{sec:abs} for hyperparameter analysis.

\xhdr{Compared methods} We compare MVGFDR with two types of DR methods: 1) Single-view methods: Inception\_ResNet~\cite{szegedy2017inception}, MobileNet~\cite{sandler2018mobilenetv2}, ResNet~\cite{he2016deep}, ResNext~\cite{xie2017aggregated}, ConvNeXt-B~\cite{liu2022convnet}, Swin-B~\cite{liu2021swin}, Vim~\cite{vim}, PVT~\cite{wang2021pyramid} and RETFound~\cite{zhou2023foundation}. 2) Multi-view methods: MVCINN~\cite{luo2023mvcinn}, ETMC~\cite{han2022trusted}, LFMVDR~\cite{luo2024lesion}, MVTSM~\cite{lin2025multi} and SMVDR~\cite{Luo_Xu_Wu_Liu_Lai_Shen_2025}. It is worth noting that LFMVDR, MVTSM, and SMVDR are not end-to-end DR methods. We compare both their expert-assisted variants (\eg, lesions or visual prompts) and their counterparts without additional information to evaluate the performance differences.

\xhdr{Evaluation metrics} For the five-grade DR grading task, we adopt commonly used evaluation metrics including accuracy (Acc.), precision (Prec.), sensitivity (Sens.), specificity (Spec.), and AUC~\cite{davis2006relationship}. Following~\cite{Luo_Xu_Wu_Liu_Lai_Shen_2025}, we also include Cohen’s kappa~\cite{cohen1960coefficient} and F$_1$ score to account for class imbalance.

\subsection{Comparison with State-of-the-art Methods}\label{sec:results}

\xhdr{Comparison with single-view methods} We report the comparison results against single-view methods on MFIDDR in Table~\ref{tab:result1}. Our proposed MVGFDR takes all data from V1-V4 views for training, while all the single-view methods separately use fours views as training data, and the view with the highest accuracy is chosen for comparison (We add ``$\_${BSV}'' to the end of each single view method to represent best single view). As can be seen from the table, our proposed MVGFDR achieves a significant performance gain against existing single-view methods (\ie, Acc. increased $11.48\%$ compared to Vim, and Spe. increased $15.69\%$ compared to PVT-L). We also report the comparison results against single-view methods on precision, sensitivity, and F$_1$ scores of different DR grades. As shown in Table~\ref{tab:per_result1}, our method consistently outperforms existing single-view methods across most evaluation metrics for each DR grade. Even for the metrics where our method does not achieve the best result, it still ranks second and remains highly competitive. These results validate that multi-view fundus imaging offers significant advantages over single-view approaches in DR grading, as each view complements others by revealing additional pathological cues, such as lesions, that may not be visible from a single perspective.

\begin{table*}[!t]
    \centering
    \caption{Comparison of precision (Prec.), sensitivity (Sens.), and F$_1$ scores between our method and multi-view approaches in different DR grades. The best results are highlighted in bold.}
    \resizebox{\linewidth}{!}{
    \begin{tabular}{l |c c c| c c c|c c c|c c c|c c c}
        \hline
        & \multicolumn{3}{c}{Grade 0} & \multicolumn{3}{c}{Grade 1} & \multicolumn{3}{c}{Grade 2} & \multicolumn{3}{c}{Grade 3} & \multicolumn{3}{c}{Grade 4}\\
        \hline
         Method & Pre. & Sens. & F$_1$ & Pre. & Sens. & F$_1$ & Pre. & Sens. & F$_1$ & Pre. & Sens. & F$_1$ & Pre. & Sens. & F$_1$\\
         \hline
         RETFound & 80.11 & 96.33 & 87.47 & 50.20 & 27.96 & 35.92 & 54.41 & 40.44 & 46.39 & 65.79 & 67.57 & 66.67 & 90.00 & 23.08 & 36.73\\
         MVCINN & 86.71 & 96.33 & 91.26 & 68.25 & 48.10 & 56.43 & 57.44 & 61.20 & 59.26 & 70.00 & 66.22 & 68.06 & 68.42 & 33.33 & 44.83\\
         ETMC & 86.79 & \textbf{97.53} & 91.82 & 73.26 & 56.38 & 63.72 & 66.41 & 47.54 & 55.41 & 64.41 & 77.03 & 70.15 & 0.12 & \textbf{90.15} & 0.87\\
         LFMVDR (with lesion) & 89.69 & 95.20 & 92.36 & 69.53 & 63.31 & 66.28 & 62.05 & 56.28 & 59.03 & 69.48 & 72.30 & 70.86 & 54.55 & 61.54 & 57.83\\
         MVTSM (with lesion)  & 89.23 & 95.65 & 92.33 & 73.86 & 54.36 & 62.63 & 61.08 & 67.76 & 64.25 & 72.67 & 73.65 & 73.15 & 64.10 & 64.10 & 64.10\\
         SMVDR (with prompts) & 93.48 & 93.55 & 93.52 & 71.15 & \textbf{72.26} & 71.70 & 60.00 & 60.66 & 60.33 & 69.41 & \textbf{79.73} & 74.21 & \textbf{99.99} & 17.95 & 30.43\\
         \hline
         \rowcolor{myLightBlue}
         \textbf{MVGFDR (Ours)} & \textbf{94.13} & 93.52 & \textbf{93.81} & \textbf{73.70} & 71.17 & \textbf{72.41} & \textbf{67.94} & \textbf{72.03} & \textbf{69.91} & \textbf{74.13} & 75.70 & \textbf{74.92} & 88.89 & 55.12 & \textbf{68.07}\\
         \hline
    \end{tabular}}
    \label{tab:per_result2}
\end{table*}

\begin{table}[!t]
    \centering
    \caption{Comparison of experimental results between multi-view approaches and our proposed method. For fair comparison, we divide existing methods into two types: End-to-end and Expert-guided. The best results are in bold font.}
    \resizebox{\linewidth}{!}{
    \begin{tabular}{l c c c c c}
        \hline
         \textbf{Method (End-to-end)} & Acc. & Spe. & Kappa & F$_1$ & AUC\\
         \hline
         RETFound~\pub{Nature23}~\cite{zhou2023foundation} & 74.06 & 73.83 & 48.44 & 70.91 & 89.04\\
         MVCINN~\pub{AAAI23}~\cite{luo2023mvcinn} & 80.10 & 83.32 & 62.45 & 78.86 & 91.07\\
         ETMC~\pub{TPAMI22}~\cite{han2022trusted} & 81.54 & 83.44 & 64.76 & 79.74 & 93.53\\
         LFMVDR (w/o lesion)~\pub{JBHI24}~\cite{luo2024lesion} & 80.38 & 85.90 & 64.04 & 79.39 & -\\
         MVTSM (w/o lesion)~\pub{PR2025}~\cite{lin2025multi}  & 71.73 & 82.76 & - & 70.93 & -\\
         SMVDR (w/o prompts)~\pub{AAAI25}~\cite{Luo_Xu_Wu_Liu_Lai_Shen_2025} & 81.64 & 89.92 & 66.94 & 81.38 & -\\
         \hline
         \textbf{Method (Expert-guided)} & & & & &\\
         \hline
         LFMVDR (with lesion)~\pub{JBHI24}~\cite{luo2024lesion} & 82.15 & 86.97 & 66.99 & 81.26 & 94.81\\
         MVTSM (with lesion)~\pub{PR2025}~\cite{lin2025multi} & 82.61 & 86.77 & 67.97 & 81.94 & 92.24\\
         SMVDR (with prompts)~\pub{AAAI25}~\cite{Luo_Xu_Wu_Liu_Lai_Shen_2025} & 84.01 & 91.30 & 71.36 & 83.69 & 95.58\\
         \hline
         MVGFDR (CGR) & 84.52 & 91.82 & 72.46 & 84.31 & 95.42\\
         \rowcolor{myLightBlue}
         \textbf{MVGFDR (Ours)} & \textbf{85.87} & \textbf{93.00} & \textbf{73.14} & \textbf{84.49} & \textbf{96.17}\\
         \hline
    \end{tabular}}
    \label{tab:result2}
\end{table}

\xhdr{Comparison with multi-view methods} We evaluate our method against existing multi-view methods. Following~\cite{Luo_Xu_Wu_Liu_Lai_Shen_2025}, we also extend the foundation model RETFound into multi-view settings. Multi-view methods are grouped into two categories: end-to-end and expert-guided. As described in Sec.~\ref{sec:exp_setting}, for expert-guided methods, we can further distinguish them between the variants with and without expert-provided information, and report their results separately. Compared to end-to-end methods, the proposed MVGFDR achieves state-of-the-art performance across all evaluation metrics. Compared to MVCINN and ETMC, instead of blindly aggregating all information from different views, MVGFDR selectively integrates view-specific information, mitigating redundant signals and improving discriminative power.

In the absence of expert guidance, existing multi-view methods exhibit suboptimal performance, as the models lack clear visual focus and struggle to accurately distinguish between different DR grades (\ie, Acc. increased $19.71\%$ compared to MVTSM (w/o lesion) and Kappa increased $9.26\%$ compared to SMVDR(w/o prompts)). When compared with expert-guided methods, despite not using lesion maps or handcrafted prompts, our end-to-end framework still outperforms them in most cases. Specifically, compared to MVTSM, which relies on lesion maps for supervision, MVGFDR achieves performance improvements of $4.00\%$, $7.18\%$, $7.61\%$, $3.11\%$ and $4.26\%$ in Acc., Spe., Kappa, F$_1$ and AUC. Furthermore, compared to the latest SOTA method SMVDR, which utilizes expert-crafted visual prompts, our method still achieves higher scores in Acc. ($+2.20\%$), Spe. ($+1.86\%$), Kappa ($+2.49\%$), and F$_1$ score ($+0.74\%$). These improvements are largely attributed to our graph-based selection and fusion strategies, which implicitly simulate expert guidance by identifying and leveraging view-specific information. Additionally, the proposed masked cross-view reconstruction module enhances the model’s ability to reason over cross-view semantics. We also report per-grade precision, sensitivity, and F$_1$ scores in Table~\ref{tab:per_result2}. Overall, the experimental results demonstrate the superiority of MVGFDR over both end-to-end and expert-guided multi-view DR methods.

Beyond the performance on MFIDDR, we further evaluate the robustness and generalization of MVGFDR on medical imaging datasets spanning different modalities. Since MFIDDR is currently the largest publicly available fundus image dataset for multi-view DR grading, we were unable to find another public dataset of the same kind. However, we obtained a private dataset DRTiD~\cite{hou2022cross} for two-field DR grading. In addition, to evaluate our method on a different imaging modality, we utilized the CheXpert dataset~\cite{irvin2019chexpert}, which contains chest X-ray images acquired from both frontal and lateral views of the same patient. We further validated our method, with results shown in Table~\ref{tab:add_exp_mod}. Since no lesions or prompts are used during training, we only compare with MVICNN, the baseline for multi-view DR grading. In addition, we include two relevant baselines per dataset. Both tables demonstrate that our method generalizes well across datasets and imaging modalities.

\begin{table}[!t]
    \scriptsize
    \centering
    \caption{Comparison of experimental results on DRTiD and CheXpert datasets.}
    \resizebox{\linewidth}{!}{
    \begin{tabular}{l c c c c}
        \hline
         \textbf{DRTiD} & Acc. & Spe. & Kappa & F$_1$ \\
         \hline
         MVCINN~\pub{AAAI23}~\cite{luo2023mvcinn} & 85.87 & 85.08 & 74.14 & 84.28\\
         PLReMix~\pub{WACV25}~\cite{liu2025plremix} & 87.94 & 88.93 & 76.70 & 87.64\\
         Qmix~\pub{TMI25}~\cite{hou2025qmix} & 89.29 & 91.15 & 80.80 & 88.31\\
         \rowcolor{myLightBlue}
         \textbf{MVGFDR (Ours)} & \textbf{92.68} & \textbf{92.74} & \textbf{83.99} & \textbf{90.71}\\
         \hline
         \textbf{CheXpert} & Acc. & Spe. & Kappa & F$_1$ \\
         \hline
         MVCINN~\pub{AAAI23}~\cite{luo2023mvcinn} & 73.04 & 79.61 & 56.90 & 71.73\\
         TMC~\pub{TPAMI22}~\cite{han2022trusted} & 68.62 & 73.53 & 51.65 & 66.46\\
         MV-HFMD~\pub{WACV24}~\cite{black2024multi} & 71.55 & 79.03 & 55.17 & 70.29\\
         \rowcolor{myLightBlue}
         \textbf{MVGFDR (Ours)} & \textbf{76.10} & \textbf{82.91} & \textbf{61.25} & \textbf{74.61}\\
         \hline
    \end{tabular}}
    \label{tab:add_exp_mod}
\end{table}

\subsection{Additional Experiments on the Generated Multi-view DR Dataset}\label{app:mvgddr}
\begin{figure}[!t]
    \centering
    \includegraphics[width=1.0\linewidth]{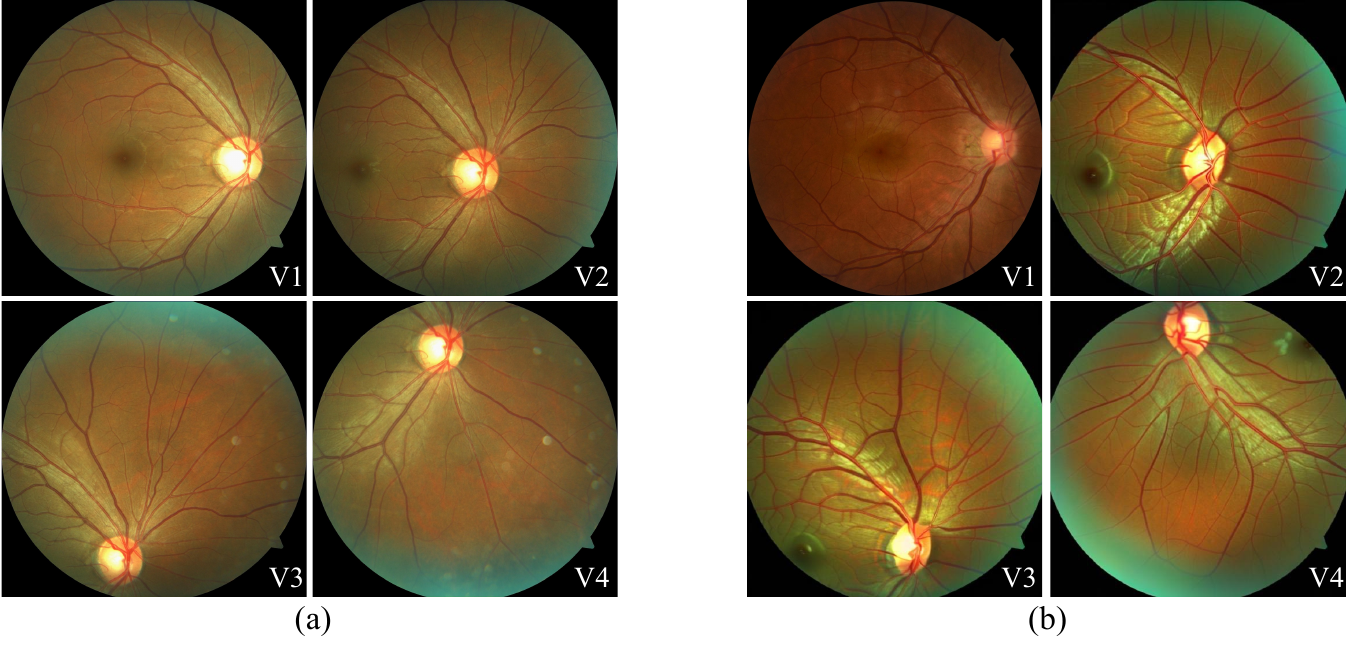}
    \caption{\textbf{Examples of V1-V4 views of different datasets.} (a) MFIDDR. (b) The generated MVG-DDR.}
    \label{fig:data_example}
\end{figure}
To further explore the generalization ability of MVGFDR, we trained a multi-view generation diffusion model~\cite{ho2020denoising} on MFIDDR and generated three additional views for the DDR~\cite{li2019diagnostic} dataset. Specifically, we adopt DreamBooth~\cite{ruiz2023dreambooth} with a pretrained Stable Diffusion model as the base generator. In each training run, we train the model to synthesize one specific view for a single DR grade. Taking Grade 3 as an example, we select a sample from Grade 3 in the DDR dataset and pair it with a V2 view image from MFIDDR. A textual prompt “\textit{a photo of v2l3sks severe diabetic retinopathy}” is used to transfer the DDR sample into the V2 style. Here, “v2l3sks” is a unique token designed to capture the style of the second view of Grade 3 in MFIDDR. To prevent the model from forgetting class-specific information, we introduce an additional L2 loss as the target function. The whole training process is conducted on two NVIDIA RTX A5000 GPUs with batch size set to 4. The model is trained for 15,000 iterations using 8-bit Adam~\cite{dettmers2022bit} optimizer with learning rate set to 2e-6. Therefore, following this training paradigm, we train 15 generation models in total for five classes (Grade 0–4) across three views (V2–V4), and applied them to the DDR dataset to generate corresponding V2–V4 samples for each input image. The newly generated dataset is referred to as \textbf{Multi-view generated DDR} (MVG-DDR). Examples from both MFIDDR and the generated MVG-DDR are provided in Fig.~\ref{fig:data_example}. The newly generated MVG-DDR is available via this \href{https://figshare.com/s/58ea812b168eb9f9d4b0}{link}\footnote{https://figshare.com/s/58ea812b168eb9f9d4b0}.
\begin{table}[!t]
    \centering
    \caption{Comparison of experimental results on the generated MVG-DDR dataset. We re-conduct experiments on single-veiw methods and multi-view methods. The best results are in bold font.}
    \resizebox{\linewidth}{!}{
    \begin{tabular}{l c c c c c}
        \hline
         \textbf{MVG-DDR (Single-view)} & Acc. & Pre. & Spe. & Kappa & F$_1$ \\
         \hline
         ResNet-50\_BSV~\pub{CVPR16}~\cite{he2016deep} & 76.59 & 74.63 & 78.80 & 65.00 & 74.12\\
         DenseNet-121\_BSV~\pub{CVPR17}~\cite{huang2017densely} & 73.57 & 71.85 & 72.58 & 60.53 & 69.54\\
         SE-BN-Inception\_BSV~\pub{CVPR18}~\cite{hu2018squeeze} & 76.39 & 74.83 & 80.94 & 64.72 & 74.62\\
         Swin-B\_BSV~\pub{CVPR21}~\cite{liu2021swin} & 78.75 & 77.09 & 79.17 & 68.88 & 76.47\\
         PVT-M\_BSV~\pub{ICCV21}~\cite{wang2021pyramid} & 79.08 & 77.72 & 78.52 & 69.08 & 76.62\\
         PVT-L\_BSV~\pub{ICCV21}~\cite{wang2021pyramid} & 80.85 & 79.74 & 81.96 & 73.43 & 79.25\\
         RETFound\_BSV~\pub{Nature23}~\cite{zhou2023foundation} & 80.10 & 79.06 & 83.90 & 72.99 & 79.15\\
         Vim\_BSV~\pub{ICML24}~\cite{vim} & 81.92 & 80.49 & 82.81 & 74.95 & 80.27\\
         \hline
         \textbf{MVG-DDR (Multi-view)} & & \\
         \hline
         RETFound~\pub{Nature23}~\cite{zhou2023foundation} & 82.52 & 80.74 & 85.11 & 76.39 & 80.86\\
         MVCINN~\pub{AAAI23}~\cite{luo2023mvcinn} & 83.45 & 81.86 & 85.59 & 78.21 & 82.02\\
         ETMC~\pub{TPAMI22}~\cite{han2022trusted} & 81.82 & 80.39 & 82.59 & 74.73 & 80.15\\
         \hline
         \textbf{MVGFDR (Ours)} & \textbf{84.87} & \textbf{83.13} & \textbf{86.78} & \textbf{79.70} & \textbf{83.52}\\
         \hline
    \end{tabular}}
    \label{tab:mvgddr}
\end{table}

\begin{figure*}[!t]
    \centering
    \includegraphics[width=1\linewidth]{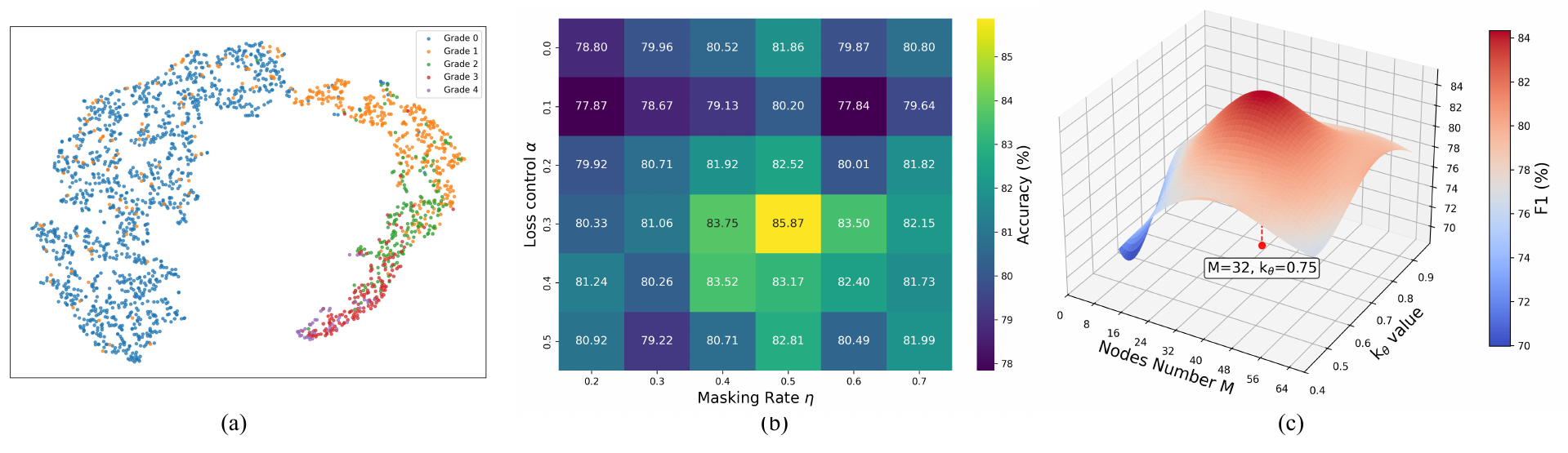}
    \caption{(a) t-SNE~\cite{van2008visualizing} visualization of feature representations learned by the model. (b-c) Evaluation of the hyper-parameter settings. (b) Joint accuracy under different masking rates $\eta$ and reconstruction loss control rate $\alpha$. (c) Joint F$_1$ score under varying numbers of clustering center $M$ and frequency range thresholds $k_\theta$. }
    \label{fig:ablation}
\end{figure*}

We also conduct experiments on MVG-DDR and compare MVGFDR with two types of DR methods: 1) Single-view methods: ResNet-50~\cite{he2016deep}, DenseNet-121~\cite{huang2017densely}, SE-BN-Inception~\cite{hu2018squeeze}, Swin-B~\cite{liu2021swin}, PVT~\cite{wang2021pyramid}, RETFound~\cite{zhou2023foundation} and VIM~\cite{vim}. 2) Multi-view methods: RETFound~\cite{zhou2023foundation}, ETMC~\cite{luo2023mvcinn} and MVCINN~\cite{luo2023mvcinn}. Following the same experimental settings as Sec.~\ref{sec:results}, all the single-view methods separately use four views of the generated MVG-DDR dataset as training data, and the view with the highest accuracy is chosen for comparison (which are shown as ``\_BSV''). Due to the lack of lesion details and visual prompts, LFMVDR, MVTSM and SMVDR are not included in the experiments.

As can be seen from Table~\ref{tab:mvgddr}, our proposed MVGFDR consistently outperforms all baseline methods, achieving notable improvements across key metrics (e.g., +1.7\% Acc., +1.5\% Pre., +1.4\% Spe., +1.9\% Kappa, and +1.8\% F$_1$ compared to MVCINN). These results highlight the strong generalization ability of MVGFDR across different datasets. Nonetheless, the performance gain on the newly generated MVG-DDR dataset is less pronounced than on MFIDDR, which we primarily attribute to the quality of the synthesized images. As illustrated in Fig.~\ref{fig:data_example}, the generated samples tend to retain stylistic characteristics of MFIDDR. We leave improving the view synthesis process to mitigate this bias as a direction for future work.

\begin{table}[t!]
	\centering
        \caption{ Ablation study of the key components on MFIDDR.}
        \resizebox{\linewidth}{!}{
		\begin{tabular}{l|ccc|ccccc}
			\cline{1-9}
			\multicolumn{1}{l}{}&\multicolumn{3}{c|}{Candidate}
			& \multicolumn{5}{c}{MFIDDR} \\
			\cline{1-9}
			\multicolumn{1}{l}{}&PVT+GF &Node Select. &View Recon. & Acc. &Pre. & Spe. & Kappa & F$_1$\\
			\hline
			\multicolumn{1}{l|}{No.1}&\ding{52}&&&79.96 &78.63 &80.17 &61.28 &77.91\\
			\multicolumn{1}{l|}{No.2}&\ding{52}&\ding{52}&&81.10 &79.87 &81.93 &63.52 &78.76\\
			\multicolumn{1}{l|}{No.3}&\ding{52}&&\ding{52}&82.52 &80.74 &85.11 & 67.39 & 80.86\\
			\multicolumn{1}{l|}{\textbf{Ours}}&\ding{52}&\ding{52}&\ding{52}&\textbf{85.87} &\textbf{85.13} &\textbf{93.00} &\textbf{73.14} &\textbf{84.49}\\
			\hline
		\end{tabular}}
        \label{tab:ablation}
\end{table}

\subsection{Ablation study}\label{sec:abs}

\xhdr{t-SNE visualization} 
We visualize the learned feature representations of MVGFDR using t-SNE, as shown in Figure\ref{fig:ablation}(a). The results demonstrate clear inter-class separation, with smooth transitions between adjacent grades. However, due to data imbalance, particularly the scarcity of Grade 4 samples and the dominance of Grade 0, the clusters of Grade 3 and Grade 4 overlap, and some Grade 1 samples are embedded within the Grade 0 cluster.

\xhdr{Effectiveness of each component} We herein validate the effectiveness of the proposed node selection strategy and masked cross-view reconstruction (MCVR) module in MVGFDR. We first compare the proposed Cross-View Reconstructor with the GCN-based baseline (MVGFDR (CGR)) introduced in Sec.~\ref{sec:gcr}. As shown in Table~\ref{tab:result2}, our decoder-only reconstructor guided by VP and FP achieves more accurate feature reconstruction compared to the simple GCN structure, demonstrating the effectiveness of incorporating view- and frequency-aware positional priors. Furthermore, we conduct additional ablation studies to analyze the contributions of each component. 
As summarized in Table~\ref{tab:ablation}, we use No.1~$\sim$~No.3 to represent different experimental settings. Specifically, No.1 serves as the baseline that adopts PVT as the backbone and directly fuses all multi-view features via graph fusion, leading to information redundancy and suboptimal performance.
To assess the effect of the proposed reconstruction, we further evaluate two variants: (No.2) without masked reconstruction and (No.3) with masked reconstruction but without node selection. Compared to No.2, our full model with MCVR achieves superior results, demonstrating that cross-view reconstruction effectively strengthens inter-view relationship modeling. However, No.3 performs worse than MVGFDR, as reconstructing all nodes indiscriminately ignores that only low- and mid-frequency components share cross-view similarity, whereas high-frequency lesion-related cues are view-specific.

\xhdr{Hyper-parameters Analysis} Lastly, we further evaluate the hyper-parameters in our approach. In Fig.~\ref{fig:ablation}(b), we provide the heatmap for the joint accuracy under different masking rates $\eta$ and reconstruction loss control rate $\alpha$. As can be seen from the figure, varying the values of $\eta$ or $\alpha$ in either direction does not lead to better performance, confirming the effectiveness of our chosen hyperparameter settings. A large $\alpha$ forces the model to overemphasize reconstruction details, while the shared information across graphs is similar but not identical. On the other hand, a large $\eta$ increases the reconstruction difficulty by masking too many nodes. We also provide the 3D surface plot in Fig.~\ref{fig:ablation}(c) for joint F$_1$ score under varying numbers of clustering centers $M$ and frequency range threshold $k_{\theta}$. As indicated by the peak location and the smooth transitions in the surrounding region, our setting strikes a balance between expressiveness and complexity, enabling sufficient representation of graph information without making the graph overly complex.

\section{Conclusion}\label{sec:5}
In this paper, we have proposed a novel end-to-end multi-view graph fusion framework MVGFDR for DR grading. Specifically, MVGFDR employs a multi-view graph initialization module to guide the graph construction process with residual as guidance and DCT coefficients served as anchors. Then, graph fusion is implemented to the high-frequency graph nodes selected by DCT frequency to enhance the model's understanding of the view-specific information. Finally, masked cross-view reconstruction is utilized to model shared patterns across views and enhance the representation of view-invariant information. Extensive experiments on the largest multi-view fundus image dataset MFIDDR have demonstrated that our MVGFDR achieves a SOTA performance against both single-view and multi-view methods.

\section*{Acknowledgments}
The authors acknowledge NVIDIA and its research support team for the help provided to conduct this work. This work was also partially supported by the Australian Research Council (ARC) Industrial Transformation Training Centres (ITTC) for Innovative Composites for the Future of Sustainable Mining Equipment under Grant IC220100028.

{\appendices
\section{Proof of Frequency-aware consistency}\label{app:dct_proof}
Although we initialize the graph construction with DCT coefficients as anchors, we treat them as learnable parameters. As described in Sec.~\ref{sec:mvgf}, our graph initialization process is designed such that, even as these parameters are updated during training, the frequency separation guided by DCT remains intact. In this section, we provide a theoretical justification for this property.

\textbf{Assumption 1.} The learnable anchor matrix $\mathbf{W} \in \mathbb{R}^{M \times C}$ is initialized from the standard DCT basis.

\textbf{Proposition 1.} Let $w_n \in \mathbf{W}$ denotes the $n$-th DCT coefficient function. Under the update rule of gradient descent, $w_n$ preserves its frequency-specific properties during training, thereby maintaining frequency separation.

\textit{Proof.} Based on Eq.~\ref{eq:DCT}, we can express $w_n$ as:
\begin{equation}
    w_n = \sum_{c=0}^{C_i - 1} a_c \cdot \phi_c(n), \text{ } \phi_c(n) = \text{cos}[\frac{\pi}{C_i}(c+\frac{1}{2})n],
\end{equation}
where $a_n$ denotes a learnable coefficient, and $\phi_c(\cdot)$ represents a fixed orthogonal DCT basis function.
As can be seen, $a_n$ only modulates the amplitude of the DCT frequency component, the resulting $w_n$ still adheres to the original frequency band structure defined by the DCT basis. Assuming the total network loss is $\mathcal{L}$, the coefficients are updated by backpropagation
\begin{equation}
    a_c^{t+1} = a_c^t - \lambda \cdot \frac{\partial \mathcal{L}}{\partial a_c},
\end{equation}
where $\lambda$ denotes the learning rate. Consequently, the updated anchor becomes
\begin{equation}
    w_n^{t+1} = \sum_{c=0}^{C_i - 1} a^{t+1}_c \cdot \phi_c(n).
\end{equation}
Due to $w_n^{(t)} \in \text{span}\{\phi_0(n), \phi_1(n), ..., \phi_{C_i-1}(n)\}$, and $\phi_c(n)$ forms a fixed orthogonal basis, it holds for any training step $t$ that
\begin{equation}
    w_n^{t} \in \mathcal{V}_{\text{DCT}} = \text{span}\{\phi_0(n), \phi_1(n), ..., \phi_{C_i-1}(n)\}.
\end{equation}
Since the updates only occur on the coefficients $a_c$ and do not alter the basis, the process essentially preserves the underlying frequency structure:
\begin{equation}
    \mathcal{V}_{\text{DCT}}^{(t+1)} =  \mathcal{V}_{\text{DCT}}^t = \mathcal{V}_{\text{DCT}},
\end{equation}
Therefore, despite gradient-based updates, the updated anchor $w_n^t$ remains projected onto the same frequency-domain subspace as the initial $w_n$,  preserving its spectral semantics.

}

{\small

\bibliographystyle{IEEEtran}
\bibliography{main}

}

\end{document}